\documentclass[]{fairmeta}
\usepackage{pgfplots}
\pgfplotsset{compat=newest}
\usepackage{tikz}

\title{Scaling A Simple Approach to Zero-Shot Speech Recognition}

\author[1,*]{Jinming Zhao}
\author[2]{Vineel Pratap}
\author[2]{Michael Auli}

\affiliation[1]{Monash University}
\affiliation[2]{Meta FAIR}

\contribution[*]{Work done during an internship at Meta FAIR.}

\abstract{
\begin{abstract}
 DDespite rapid progress in increasing the language coverage of automatic speech recognition, the field is still far from covering all languages with a known writing script.
Recent work showed promising results with a zero-shot approach requiring only a small amount of text data, however, accuracy heavily depends on the quality of the used phonemizer which is often weak for unseen languages.
In this paper, we present MMS Zero-shot a conceptually simpler approach based on romanization and an acoustic model trained on data in 1,078 different languages or three orders of magnitude more than prior art.
MMS Zero-shot reduces the average character error rate by a relative 46\% over 100 unseen languages compared to the best previous work.
Moreover, the error rate of our approach is only 2.5x higher compared to in-domain supervised baselines, while our approach uses no labeled data for the evaluation languages at all.
Code and models are available at \url{https://github.com/facebookresearch/fairseq/tree/main/examples/mms/zero_shot}
\end{abstract}}

\date{\today}
\correspondence{\email{jinming.zhao@monash.edu}, \email{vineelkpratap@meta.com}, \email{michaelauli@meta.com}}

\begin{document}

\maketitle

\section{Introduction}

\label{sec:introduction}

There has been significant work in enabling automatic speech recognition (ASR) for more of the over 7,000 languages spoken around the world~\citep{campbell2008ethnologue}.
One approach has been to perform self-supervised learning~\citep{oord2018cpc,baevski2020wav2vec} followed by fine-tuning on labeled data to build models supporting between 100 and 1,000 languages~\citep{zhang2023usm,pratap2024scaling}. 
Another line of work focuses on traditional supervised learning using large amounts of labeled data~\citep{radford2023robust} which is only available for a small subset of languages.
However, despite rapid progress in language expansion, it appears unlikely that it is possible to obtain a reasonable amount of labeled data for all languages with a writing script.
An alternative is to do away with labeled data altogether via unsupervised ASR methods~\citep{liu2018completely,chen2019completely,baevski2021unsup} but a drawback of these methods is the requirement for both unlabeled audio and unlabeled text which may still be difficult to obtain for many new languages.

More recently, a promising zero-shot approach has been introduced which requires only unlabeled text~\citep{li2022asr2k}.
The idea is to train an acoustic model which outputs language-independent allophones that are then mapped to language-specific phonemes~\citep{li2020universal}. 
Next, a pronunciation model utilizing a zero-shot G2P system~\citep{li2022zero} maps phoneme sequences to words and decoding may be further improved via an n-gram language model. 
This approach has the downside that the mapping from allophones to phonemes is error-prone~\citep{mortensen2020allovera} and that the G2P phonemizer has poor performance for many unseen languages as we show.\footnote{This is likely because the dataset used by the phonemizer covers only 40 out of 110 branches of the Glottolog Phylogenetic tree~\citep{li2022zero}.} 

\begin{figure}[t]
\centering
\includegraphics[width=0.55\linewidth]{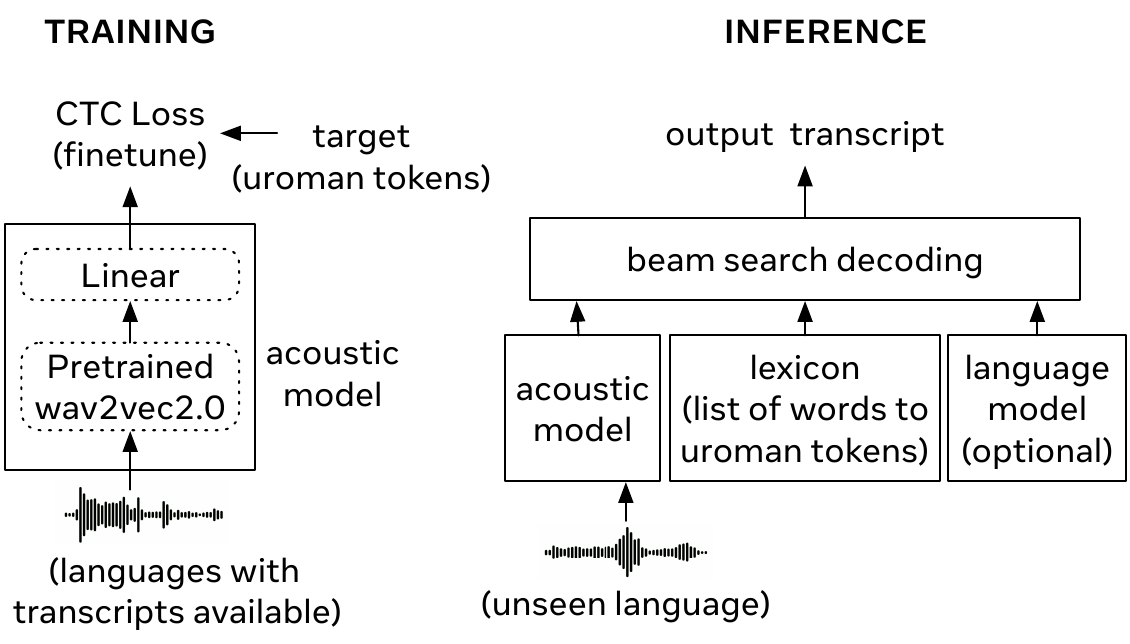}
\caption{MMS Zero-shot. We build a universal acoustic model by fine-tuning a pre-trained wav2vec 2.0 model on romanized transcripts (left). 
A new language is transcribed by performing beam search decoding with a lexicon mapping words in the new language to romanized text.
If available, then a language model can be used to improve performance (right).
}
\label{fig:zeroshot_arch}
\end{figure}

To sidestep these challenges, we do away with allophones and phonemes and use a single intermediate text representation by romanizing text in different languages which standardizes them to a common Latin-script~\citep{hermjakob2018out}.
Our acoustic model is trained on labeled data in 1,078 different languages and outputs romanized text. 
For inference, we map model outputs to words by employing a simple lexicon based on a romanized encoding of a modest amount of supplied text in a new language~(\cite{hermjakob2018out}; see Figure~\ref{fig:zeroshot_arch}).

We do not require language independent phonemizers which we find to result in poor accuracy for many languages and experiments show that this simple approach can lead to large accuracy improvements on 100 unseen MMS-lab and FLEURS languages~\citep{conneau2023fleurs,pratap2024scaling}.

\section{Method}

\subsection{Universal Acoustic Model}
\label{sec:uroman}

There are a variety of different writing scripts and we standardize different languages using romanization which essentially converts text to a single writing script~\citep{wellisch1978conversion}. We use uroman~\citep{hermjakob2018out}, a universal uromanizer, which performs the mapping using a set of heuristics, rather than by relying on language-specific dictionaries. After converting all the transcripts to their romanized version, we finetune a pretrained wav2vec 2.0~\citep{baevski2020wav2vec} model on many languages for which transcripts are available, so that the model can generalize to unseen languages. 

Universal phonemizers~\citep{li2022asr2k} are another way to standardize text, however, they require complex linguistic rules and are less robust compared to uroman, hindering their applicability to many unseen languages.\footnote{Universal phonemizers failed on certain languages of this study unlike uroman.}

\subsection{Zero-shot decoding}
\label{sec:decoding}

Zero-shot transcription requires only a list of words in the unseen language.
The first step is to produce a lexicon by applying uroman to each word so as to have \texttt{<word, uroman\_text>} pairs. 
When decoding with the lexicon, the model is forced to produce words contained in the lexicon. 
Open-source databases, such as Panlex~\citep{kamholz2014panlex} provide word lists for over 6k languages, which enables building speech recognizers for all these languages. 

We can additionally integrate a simple n-gram language model during decoding to further improve performance~\citep{pratap2019wav2letter}.
Leveraging databases with word statistics is also an alternative,
e.g., following~\citet{li2022asr2k}, we investigate the use of Crúbadán~\citep{scannell12007crubadan} which is a database that includes unigrams, bigrams, and character-level trigrams, together with their statistics, for about 2k languages.\footnote{The Crúbadán dataset was received directly from the authors in~\citet{scannell12007crubadan}. 
Meta did not scrape or collect the data from the original sources by scripts or other means.
}

\section{Experiments}

\subsection{Datasets} 
\label{sec:datasets}

We use the following datasets: (i) \textit{MMS-lab (MMS)}, built on the New Testament, has paired labeled data for 1,107 languages~\citep{pratap2024scaling},
(ii) \textit{FLEURS} is a 102-language open-sourced ASR benchmark dataset~\citep{conneau2023fleurs},
(iii) \textit{CommonVoice (CV)} is a multilingual ASR dataset, crowdsourced from global volunteers~\citep{ardila2020common} comprising 85 languages from version 8.0.
We hold out 90 randomly selected low-resource languages from MMS and 10 from FLEURS of which half are used for development and half for test.
All CV languages are used for training as most of them are high-resource.
The held-out languages are not present in the training data.



    


\subsection{Baselines}
We compared our system to results obtained with the public model checkpoint of ASR-2K, a zeroshot ASR model that supports nearly 2k languages~\citep{li2022asr2k}. 
Since their method is modular, we use our decoding strategy based on beam search, for a more like for like comparison.
We also compare to supervised models to get a sense of how much the accuracy of zero-shot systems differs from the ideal setting where labeled data is available:
We train three different sets of monolingual models for the 100 dev and test languages based on character outputs (mono-char), uroman outputs (mono-uroman), as well as phoneme outputs (mono-phone) using the universal phonemizer~\citep{li2022zero} utilized in ASR-2K.

\subsection{Training and Decoding Setup}
We fine-tune XLS-R~\citep{conneau2020unsupervised}\footnote{We do not use MMS pretrained models as the held-out languages are seen during the pretraining.} on paired speech and uroman text data using CTC~\citep{graves2006connectionist} and follow the settings of XLS-R 300M.\footnote{\url{https://github.com/facebookresearch/fairseq/blob/main/examples/wav2vec/xlsr/README.md}}
We first apply uroman to the text portion of the training data of each language which results in a shared vocabulary of less than 30 characters over all languages that we use for training.
We then fine-tune XLS-R using labeled training data from MMS and CV, comprising 1,078 languages and a total of 39,946 hours of data. 
The amount of data per language varies by language, ranging from a few minutes for low-resource languages to a thousand hours for high-resource languages.

Monolingual models are trained on 4 GPUs for 10k updates and multilingual models on 32 GPUs for 200k updates.
The decoding beam size is 2k unless otherwise mentioned. 
Models are selcted based on dev set accuracy of the training languages for supervised toplines, and on the dev sets of 50 unseen languages for zero-shot models.
We used flashlight for decoding~\citep{kahn2022flashlight} and report CER on the dev sets and test sets for our unseen dev and test languages, respectively. 
We tuned wordscore and language model weights for all models, including ASR-2K, on the dev set of 50 unseen languages, and apply the best aggregated scores to unseen languages at inference time.\footnote{
We found wordscore tuning to be crucial in obtaining good performance and that a wordscore tuned on the dev languages is very competitive to a wordscore tuned on the actual unseen languages.}

\section{Results and Analysis}

\begin{table*}[t]
\centering
\setlength{\tabcolsep}{4pt} 
\scalebox{0.85} {
\begin{tabular}{@{}lccccccccccc|cc@{}}
\toprule
& \multirow{2}{*}{\#Train} & \multicolumn{2}{c}{MMS dev} & \multicolumn{2}{c}{MMS test} &  \multicolumn{2}{c}{FLEURS dev} & \multicolumn{2}{c}{FLEURS test} & \multicolumn{2}{c}{CV test} & \multicolumn{2}{|c}{AVG} \\
\cmidrule(r){3-4} \cmidrule(lr){5-6} \cmidrule(lr){7-8} \cmidrule(lr){9-10} \cmidrule(l){11-12} \cmidrule(l){13-14}
Model & lang & lex& 1grm & lex& 1grm & lex& 1grm & lex& 1grm & lex& 1grm & lex& 1grm\\
\midrule
ASR-2K            & 8 & 47.2 & 38.2 & 50.4 & 39.5 & 55.5 & 51.4 & 62.2 & 56.1 & 45.4 & 38.7 & 52.2 & 44.8\\
MMS Zero-shot (CV-only) & 8 & 32.3 & 29.8 & 36.4 & 33.5 &48.7 & 46.1& 50.0 &46.8 & 36.0 & 34.2 & 40.7 & 38.5\\
MMS Zero-shot & 1,078 & 17.5 & 11.6 & 20.0 & 12.6 & 40.4 & 38.6 & 41.4 & 38.2 & 26.8 & 25.2 & 29.2& 25.2 \\
\bottomrule
\end{tabular}
}
\caption{Comparison to ASR-2K. We report average CER on a total of 107 unseen languages from MMS-lab, FLEURS and CV languages (\textsection\ref{sec:datasets}). The average CER is unweighted. 
}
\label{tab:ASR-2K}
\end{table*}

\begin{table*}[t!] 
\centering
\setlength{\tabcolsep}{4pt} 
\scalebox{0.85}
{

\begin{tabular}{@{}lcccccccccccc|ccc@{}}
\toprule
& \multicolumn{3}{c}{MMS dev} & \multicolumn{3}{c}{MMS test } & \multicolumn{3}{c}{FLEURS dev} & \multicolumn{3}{c}{FLEURS test }& \multicolumn{3}{|c}{AVG } \\
\cmidrule(lr){2-4} \cmidrule(lr){5-7} \cmidrule(lr){8-10} \cmidrule(lr){11-13} \cmidrule(lr){14-16}
models & lex & 1grm & 3grm  & lex & 1grm & 3grm  & lex & 1grm & 3grm  & lex & 1grm & 3grm  & lex & 1grm & 3grm \\
\midrule
mono-char &  1.9 & 1.9 & 1.8 & 2.3 & 2.3 & 2.3 & 16.8 & 17.0 & 16.9 & 16.7 & 16.8 & 16.4 &9.4 & 9.5 & 9.4 \\
mono-uroman &  5.4 & 3.1 & 2.7 & 6.0 & 3.2 & 2.9 & 18.5 & 17.7 & 17.7 & 25.4 & 20.8 & 19.6 &13.8 &11.1 & 10.7 \\
MMS Zero-shot &  17.5 & 11.6 & 9.3 & 20.0 & 12.6 & 10.7 & 40.4 & 38.6 & 38.2 & 41.4 & 38.2 & 37.8 & 29.8 & 25.3 & 24.0 \\
\bottomrule
\end{tabular}}
\caption{
Comparison of supervised models to MMS Zero-shot on 90 MMS languages and 10 FLEURS languages.
Each supervised monolingual model is trained on the domain it is evaluated on.
The average CER is unweighted.
}
\label{tab:main}
\end{table*}

\subsection{Comparison to Prior Work}
\label{sec:results_priorwork}

We first present a comparison to the most relevant prior work, ASR-2K, on three different sets of languages: 
MMS-lab (90 languages in total), FLEURS (10 languages), and 7 CommonVoice languages from CV 17 which are unseen for all models since models were trained on CV 6.\footnote{We could not exactly determine the CV version used by~\citep{li2022asr2k} and found CV 6 to be a good match of their data description.}
The average amount of target language text data to build lexicons or LMs is 9k utterances for MMS-lab, 3k for FLEURS and 6.5k for CV.

We compare three models: 
first, ASR-2K using flashlight beam search decoding, with a lexicon mapping from words to phonemes. 
Second, our uroman zero-shot approach with an acoustic model trained on the same eight languages as ASR-2K using CV 6.0 data to isolate the effect of more training languages compared to ASR-2K (MMS Zero-shot CV-only). 
Third, our uroman zero-shot approach with an acoustic model trained on all 1,078 languages of MMS-lab and CV (MMS Zero-shot). 
Models are evaluated using only a lexicon or an additional unigram language model estimated on the transcriptions of the training data of a particular language in the respective corpus.

Table~\ref{tab:ASR-2K} shows that the MMS Zero-Shot (CV-only) model reduces the error rate by between 16-22\% relative compared to ASR-2K across all 107 languages, including CV languages, which is in-domain for ASR-2K.\footnote{We select models based on unseen dev language accuracy which performs better than using seen languages.
We could not determine the ASR-2K model selection strategy, however, even if we select models based on seen languages, CER still decreases by a relative 5-15\% on average compared to ASR-2K.}
Therefore, a romanization based encoding can perform very well compared to the allophone-based approach of ASR-2K.
Moreover, our approach has no intermediate phone representations and directly predicts uroman symbols instead of requiring a learned mapping from phones to phonemes.

When our acoustic model is trained on three orders of magnitude more languages, then performance improves further: MMS Zero-shot reduces the error rate by a relative 46\% compared to ASR-2K on average across all languages. 
This is partially because the universal phonemizer of ASR-2K leads to poor accuracy on some languages (CER $\ge$ 70). 
However, even after removing these languages, MMS Zero-shot (CV-only) still outperforms ASR-2K by an average absolute CER of 9.7\%/5.1\% for lexicon/1-gram decoding. 

Moreover, we can see that unigram language models perform much better than unweighted lexicons which provide no guidance to which words a uroman sequence should be mapped when there is ambiguity.
Our zero-shot models were trained on MMS-lab data which partly explains why improvements are larger on unseen MMS-lab evaluation languages, however, there is still a very sizeable improvement on FLEURS which is out-of-domain for the uroman-based models.

\subsection{Comparison to Supervised Models}

How well does zero-shot ASR perform compared to supervised systems?
Zero-shot models are at an inherent disadvantage due to the lack of supervision in the target language but the size of the gap will help us better understand the progress in zero-shot methods.
To get a better sense of this, we compare the MMS Zero-shot model to monolingual supervised systems outputting either uroman or characters (mono-uroman, mono-char). 
The monolingual models are trained on MMS-lab and FLEURS data and we evaluate on the same benchmarks. 
This is an in-domain setting for the supervised models and presents a very strong baseline.

Table \ref{tab:main} shows CER with lexicon-based and language model decoding.\footnote{Lexicons decrease accuracy for mono-char models but we use them for all settings for a like for like comparison.} 
Language models substantially improve performance over lexicon-only decoding for uroman models, both supervised and zero-shot, since they can mitigate uroman to character mapping ambiguities.\footnote{For instance, the letter `a' can be represented with diacritical marks (e.g., `\'{a}', `\^{a}', `\r{a}'), which are all mapped to `a' during romanization which in turn leads to ambiguity during decoding.
}
The gap between MMS Zero-shot and the supervised systems varies depending on the benchmark: 
on MMS-lab languages, it is relatively large which we suspect is due to the narrow domain of the data that is based on biblical texts, and supervised models perform over-proportionally well due to the relative simplicity of the setting.
However, the gap shrinks for FLEURS whose domain is not as constrained as MMS-lab.

\begin{table}[t] 
\centering
\scalebox{0.85}{
\begin{tabular}{@{}lcccccccc}
\toprule
&  \multicolumn{2}{c}{MMS dev} & \multicolumn{2}{c}{MMS test } & \multicolumn{2}{c}{FLEURS dev  } & \multicolumn{2}{c}{FLEURS test  } \\
\cmidrule(lr){2-3} \cmidrule(lr){4-5} \cmidrule(lr){6-7} \cmidrule(l){8-9} 
 & lex & 1grm & lex & 1grm  & lex & 1grm & lex & 1grm \\
\midrule
In-domain & 16.9 & 10.4 & 19.3 & 11.7 & 39.6 & 37.6 & 41.4 & 38.2 \\
Crúbadán & 17.7 & 14.4 & 22.7 & 18.3 & 38.7 & 38.4 & 42.2 & 40.8 \\
Panlex & 51.4 & - & 52.3 & - & 57.7 & - & 52.8 & - \\
\bottomrule
\end{tabular}
} 
\caption{Comparison of Crúbadán and Panlex text resources to in-domain text data for MMS-lab/FLEURS.
We report CER on 71 (30/32/4/5) MMS-lab dev/test and FLEURS dev/test languages common to both text resources.
}
\label{tab:database}
\end{table}

\begin{table*}[h!]
\centering
\scalebox{0.85}{
\begin{tabular}{@{}lcccccccc|cc@{}}
\toprule

 & \multicolumn{2}{c}{MMS dev } & \multicolumn{2}{c}{MMS test } & \multicolumn{2}{c}{FLEURS dev } & \multicolumn{2}{c}{FLEURS test} & \multicolumn{2}{|c}{AVG}\\
\cmidrule(r){2-3} \cmidrule(lr){4-5} \cmidrule(lr){6-7} \cmidrule(l){8-9} \cmidrule(l){10-11}
Model & lex & 1grm & lex & 1grm & lex & 1grm & lex & 1grm & lex & 1grm \\
\midrule
mono-phone  & 15.3 & 7.7 & 14.4 & 5.8 & 54.0 & 49.0 & 53.4 & 41.0 & 34.3 & 27.6 \\
mono-uroman & 5.8  & 3.1 & 6.8  & 3.5 & 18.5 & 17.7 & 25.4 & 20.8 & 14.1 & 11.9\\
\bottomrule
\end{tabular}
}
\caption{
Comparison between text representations for acoustic models. 
A language-independent phonemizer (phone) performs less well than uroman transliterations.
}

\label{tab:phone}
\end{table*}

\begin{table}[h!]
\centering
\scalebox{0.85}{
\begin{tabular}{lcc}
\toprule
Model                     & lex& 1grm \\ 
\midrule 
mono-uroman                     & 8.7          & 4.6            \\
mono-phone                     & 17.2         & 10.3           \\
mono-phone (lang-specific) & 3.2   & 2.0            \\ 
\bottomrule
\end{tabular}
}
\caption{Comparison between a uroman text encoding, a language-independent phonemizer (phone) and language-specific phonemizers to process text for eight MMS-lab languages.}
\label{tab:epitran}
\end{table}

The difference between mono-char and mono-uroman shows that using romanization as text representation is inferior to the actual characters of each language but the drop is relatively small at 1.3\% CER absolute when averaged over the four settings. 
Overall, the CER of MMS Zero-shot is on average 2.5 times higher compared to character-based monolingual systems, however, the latter represent the ideal setting where in-domain labeled data is available and systems are focused on a single language.
We regard this as a very encouraging result, given the advantage the supervised systems in this comparison enjoy.

\subsection{Leveraging Existing Low-Resource Text Data}

The zero-shot approach requires only text data in the unseen languages, however, obtaining such data can be challenging for less spoken languages.
To get a better sense of how well our approach performs, we evaluate it using the Crúbadán~\cite{scannell12007crubadan} and Panlex~\cite{kamholz2014panlex} low-resource text databases which enable constructing lexicons and even simple unigram language models similar to ASR-2K~\citep{li2022asr2k} for very low resource languages.

Table~\ref{tab:database} shows results for decoding with lexicons and language models on these databases for languages covered by these resources and compares to in-domain text data as a lower bound.
Crúbadán performs competitively to the in-domain setup with lexicon-only decoding for both MMS-lab and FLEURS.
In-domain unigram language models substantially improve over Crúbadán for MMS-lab but less so for FLEURS. 
This may be due to the larger amount of MMS-lab data being available and MMS-lab being easier to transcribe because it is from a relatively narrow domain.

Panlex has much broader language coverage but performs less well due to the much smaller amounts of data: 
for the approximately 1,700 languages covered by both databases, the median number of words per language in Panlex is 191 words compared to 4,670 for Crúbadán.
Moreover, Panlex contains very few words for many languages which makes it virtually impossible to use for many low resource languages.
It also contains a large amount of noise, and a handful of languages have incorrect scripts.

\subsection{Amount of Text Data}

To get a sense of how much text data to collect for a new language, we measure the accuracy of zero-shot decoding when using increasing amounts of text data in order to build lexicons and unigram language models.
Data is drawn from CommonCrawl~(CC; \citep{heafield2011kenlm,conneau2020unsupervised}), a web domain corpus, and we measure accuracy on 5 FLEURS dev languages which are low-resource.
We compare this to the ideal setting where we build lexicons and unigram LMs based on approximately 3k utterances of in-domain FLEURS text data (FLEURS-lex/1gram topline).

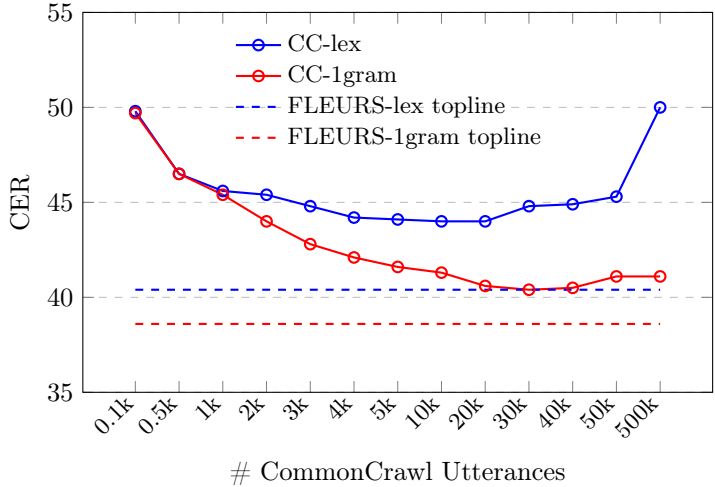
\begin{figure}
    
    \centering 
    \begin{tikzpicture} [trim axis left]
    \begin{axis}[
        width=.6\textwidth,
        height=.4\textwidth,
        xlabel={\# CommonCrawl Utterances},
        ylabel={CER},
        xtick=data,
        xticklabels={0.1k, 0.5k, 1k, 2k, 3k, 4k, 5k, 10k, 20k, 30k, 40k, 50k, 500k},
        x tick label style={rotate=45, anchor=east, font=\fontsize{9}{10}\selectfont},
        ymin=35, ymax=55,
        ymajorgrids=true,
        grid style=dashed,
        legend style={font=\footnotesize\small, legend cell align=left, at={(0.75,0.97)}, draw=none, fill=none},
        cycle list name=color list
    ]
    
    \addplot[
        color=blue,
        mark=o,
        thick
    ] coordinates {
        (0, 49.8) (1, 46.5) (2, 45.6) (3, 45.4) (4, 44.8) (5, 44.2) (6, 44.1) (7, 44.0) (8, 44.0) (9, 44.8) (10, 44.9) (11, 45.3) (12, 50.0)
    };
    \addlegendentry{CC-lex}

    \addplot[
        color=red,
        mark=o,
        thick
    ] coordinates {
        (0, 49.7) (1, 46.5) (2, 45.4) (3, 44.0) (4, 42.8) (5, 42.1) (6, 41.6) (7, 41.3) (8, 40.6) (9, 40.4) (10, 40.5) (11, 41.1) (12, 41.1)
    };
    \addlegendentry{CC-1gram}
    
    \addplot[dashed, blue, thick] coordinates {(0,40.4) (12,40.4)};
    \addlegendentry{FLEURS-lex topline}

    \addplot[dashed, red, thick] coordinates {(0,38.6) (12,38.6)};
    \addlegendentry{FLEURS-1gram topline}
    
    \end{axis}
\end{tikzpicture}
\caption{
Accuracy on FLEURS dev languages when increasing the amount of CommonCrawl text data to build lexicons and unigram LMs compared to using in-domain text data (FLEURS-topline; 3k utterances). 
\label{fig:cc}
}
\end{figure}

Figure~\ref{fig:cc} shows that having 5k utterances would allow us to build a lexicon resulting in an average CER of 44.2, versus 40.4 with the in-domain lexicon. 
More utterances does not necessarily lead to better results due to the many-to-one correspondence issue with uroman text discussed earlier. 
The gap can be further reduced by using out-of-domain language models: 
unigram models based on 20k utterances reduce CER to 40.5, compared to 38.6 CER with the FLEURS in-domain language models, however, gains plateau with more data.

\subsection{Ablation of Text Representation}

A key difference between ASR-2K and our approach is the text representation of the acoustic model.
To provide a controlled like for like comparison between the text representations, we train supervised monolingual models outputting either phonemes (mono-phone) or romanized (mono-uroman) text on eight mid- to high-resource MMS-lab languages.\footnote{These languages are eng, hin, vie, nya, tgl, ara, spa and khm.}
The phoneme models are trained on transcriptions processed by a multilingual phonemizer~\citep{li2022zero} and use lexicons mapping words to phoneme sequences.

The results (Table~\ref{tab:phone}) show that mono-uroman models perform much better than mono-phone without any language specific representation across all settings, reducing the average CER by a relative 57-59\%.
The performance difference is largely due to the phonemizer not performing well on many unseen languages~\citep{li2022zero}. 
When we switch to language-specific phonemizers\footnote{\url{https://github.com/dmort27/epitran}} then the performance vastly improves (Table~\ref{tab:epitran}), however, access to a language-specific phonemizer for unseen languages is highly unlikely.

\section{Conclusion}

We present an improved zero-shot approach to automatic speech recognition which is based on uroman transliteration and an acoustic model trained on three orders of magnitude more languages, both of which yield substantial improvements over prior art.
The approach requires only a moderate amount of text data to enable ASR for unseen languages and reduces the character error rate on average by a relative 46\% over ASR-2K on 100 languages.
Compared to supervised systems using trigram language models, the zero-shot approach produces error rates which are only about 2.5x higher.

\clearpage
\newpage
\bibliographystyle{assets/plainnat}
\bibliography{paper}

\clearpage


\end{document}